\documentclass[final,5p,times,twocolumn,authoryear]{elsarticle}

\usepackage{amssymb}
\usepackage{amsmath}
\usepackage{multirow}
\usepackage{booktabs}
\usepackage{tabularx}
\usepackage{pifont}
\usepackage{url}
\usepackage[dvipsnames]{xcolor}

\newcommand{\cmark}{\ding{51}}%
\newcolumntype{M}[1]{>{\centering\arraybackslash}m{#1}}

\journal{Expert Systems With Applications}

\begin{document}

\begin{frontmatter}




\title{Optimized Information Flow for Transformer Tracking}

\author[1,2]{Janani Kugarajeevan}
\ead{jananitha@univ.jfn.ac.lk}

\author[3]{Thanikasalam Kokul\corref{cor1}}
\ead{kokul@univ.jfn.ac.lk}

\author[3]{Amirthalingam Ramanan}
\ead{a.ramanan@univ.jfn.ac.lk}

\author[1]{Subha Fernando}
\ead{subhaf@uom.lk}

\cortext[cor1]{Corresponding author}

\affiliation[1]{organization={Department of Computational Mathematics, University of Moratuwa},
	country={Sri Lanka}}
\affiliation[2]{organization={Department of Interdisciplinary Studies, University of Jaffna},
	country={Sri Lanka}}
\affiliation[3]{organization={Department of Computer Science, University of Jaffna},
	country={Sri Lanka}}

\begin{abstract}
One-stream Transformer trackers have shown outstanding performance in challenging benchmark datasets over the last three years, as they enable interaction between the target template and search region tokens to extract target-oriented features with mutual guidance. Previous approaches allow free bidirectional information flow between template and search tokens without investigating their influence on the tracker's discriminative capability. In this study,  we conducted a detailed study on the information flow of the tokens and based on the findings, we propose a novel Optimized Information Flow Tracking (OIFTrack) framework to enhance the discriminative capability of the tracker. The proposed OIFTrack blocks the interaction from all search tokens to target template tokens in early encoder layers, as the large number of non-target tokens in the search region diminishes the importance of target-specific features. In the deeper encoder layers of the proposed tracker, search tokens are partitioned into target search tokens and non-target search tokens, allowing bidirectional flow from target search tokens to template tokens to capture the appearance changes of the target. In addition, since the proposed tracker incorporates dynamic background cues, distractor objects are successfully avoided by capturing the surrounding information of the target. The OIFTrack demonstrated outstanding performance in challenging benchmarks, particularly excelling in the one-shot tracking benchmark GOT-10k, achieving an average overlap of 74.6\%. The code, models, and results of this work are available at \url{https://github.com/JananiKugaa/OIFTrack}.
\end{abstract}


\end{frontmatter}

\section{Introduction}\label{sec1}

Visual Object Tracking (VOT) involves estimating the location and size of an object from the initial frame of a video sequence and persistently tracking it across successive frames.  A wide range of applications relies on VOT across various domains, including video surveillance \citep{Choubisa_2023_Object},  automated driving \citep{Bai_2022_Infrastructure}, augmented reality \citep{Baker2023Localization}, mobile robotics \citep{ZHANG2023Automated}, traffic monitoring \citep{Yang2023Cooperative}, and human-computer interaction \citep{Cao2023Eye}. Although many single VOT approaches have been proposed in the last three decades, achieving tracking robustness at a human-level standard remains a significant challenge due to various complicating factors encountered in real-world scenarios, such as appearance variations, occlusions, motion blurring, background clutter, and the presence of similar object distractors.

In the last decade, deep learning-based approaches, particularly those utilizing Convolutional Neural Network (CNN) based tracking methods, have demonstrated outstanding performance. Specifically, Siamese-based CNN tracking methods \citep{li2019siamrpn++,chen2020siamese, zhang2020ocean, danelljan2020probabilistic, voigtlaender2020siam, guo2021graph, mayer2021learning, thanikasalam2019target, Liang_Li_Long_2023} have shown excellent performance in appearance-based single-object tracking. Siamese-based trackers employ two identical CNN branches \citep{krizhevsky2017imagenet} to independently extract features from a target template and a search region, then locate the target within the search region by computing the similarity between the target and search region features using a simple cross-correlation operation. Although Siamese-based trackers have successfully balanced tracking accuracy and efficiency, their performance is considerably limited due to the neglect of the global context in feature extraction by CNNs. Furthermore, the correlation operation is inadequate for capturing non-linear interactions, such as occlusion, deformation, and rotation between the target template and search region, thereby limiting the tracking performance of Siamese-based trackers. Due to the limitations of CNNs in VOT, a few recent tracking approaches \citep{kugarajeevan2023transformers} have employed Transformers in single-object tracking.

Transformer \citep{vaswani2017attention}, initially introduced in the field of Natural Language Processing (NLP) and became popular due to its parallelization capabilities, scalability to handle long sequences, and effectiveness in capturing contextual information through attention mechanisms. Inspired by the success of Transformers in NLP tasks, researchers have modified \citep{dosovitskiy2021image} the architecture for vision tasks, demonstrating outstanding performance in various domains such as object detection \citep{liu2023continual}, image generation \citep{kim2023magvlt},  image classification \citep{zhou2023feature}, semantic segmentation \citep{shi2023transformer}, and point cloud comprehension \citep{yu2023rotation}. Over the past three years, owing to the tremendous success of Transformers in various computer vision tasks, researchers have developed a set of Transformer-based VOT approaches \citep{kugarajeevan2023transformers}, demonstrating excellent performance on benchmark datasets.

Transformer-based tracking approaches \citep{Chen2023seqtrack, Wei2023autoregressive, gao2023generalized, xie2023videotrack,  He_Zhang_Xie_Li_Wang_2023, Cai_2023_ICCV, Song_Luo_Yu_Chen_Yang_2023, Gopal_2024_WACV, Yang_2023_IEEETrans, Kang_2023_ICCV, Zhao_2023_CVPR, ye2022joint, chen2022backbone,cui2022mixformer,lin2022swintrack,fu2022sparsett,gao2022aiatrack,song2022transformer,yu2021high,yan2021learning, chen2021transformer} have shown better tracking accuracies than the CNN-based trackers in the last three years. In some early approaches, Transformers were utilized as a correlation module alongside the CNN backbone within a two-stream pipeline, combining both CNN and Transformer architectures \citep{gao2022aiatrack,song2022transformer,yu2021high,yan2021learning, chen2021transformer} for feature extraction and relation modeling. Since these hybrid CNN-Transformer VOT approaches still depend on a CNN for feature extraction, they are unable to capture a global feature representation of a target object. To address this limitation, some approaches \citep{lin2022swintrack,fu2022sparsett} completely replace the CNN backbone with the Transformer in a two-stream pipeline, extracting features of the template and search region using two individual Transformer models. However, as these two-stream Transformer trackers fail to consider the information flow between the template and search region during feature extraction, their tracking abilities are considerably limited. To address this limitation, more recently, researchers have introduced one-stream Transformer-based approaches \citep{Chen2023seqtrack, Wei2023autoregressive, gao2023generalized, xie2023videotrack, He_Zhang_Xie_Li_Wang_2023, Cai_2023_ICCV, Song_Luo_Yu_Chen_Yang_2023, Kang_2023_ICCV, Zhao_2023_CVPR, cui2022mixformer, ye2022joint, chen2022backbone} in VOT, combining feature extraction and relation modeling into a unified approach using a single Transformer model. A recent study \citep{kugarajeevan2023transformers} showed that one-stream Transformer-based approaches outperformed CNN-based, CNN-Transformer, and two-stream Transformer-based approaches by a large margin in terms of tracking accuracies on all challenging benchmark datasets.  

\begin{figure}[t]%
	\centering
	\includegraphics[width=0.48\textwidth]{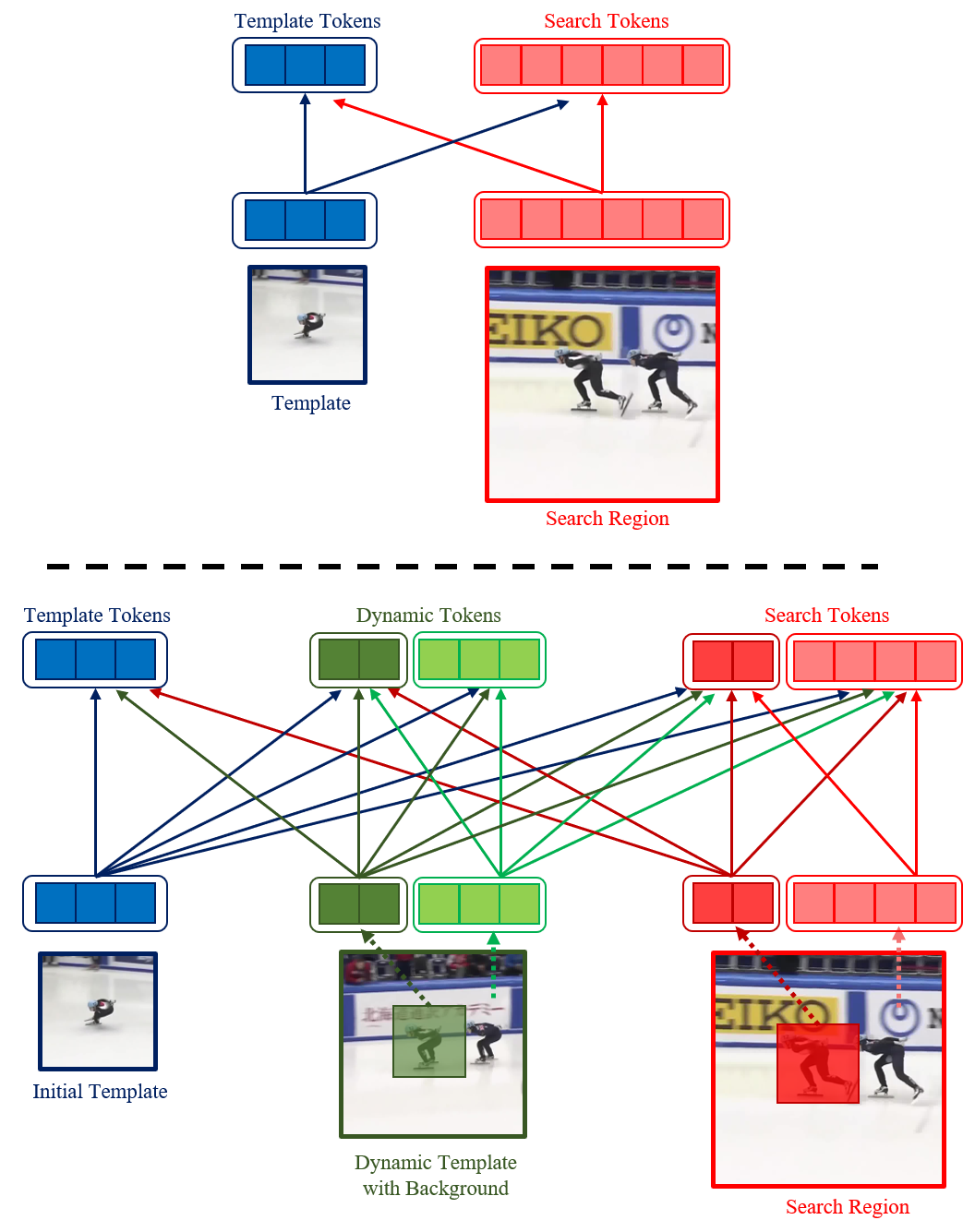}
	\caption{Comparison of the information flow of tokens between previous one-stream Transformer trackers (top) and proposed approach (bottom). Attention between all target and search region tokens are computed in previous one-stream Transformer trackers, while the proposed approach selectively blocks attention to enhance the discriminative ability of the tracker.  }
	\label{fig:introduction}
\end{figure}

In the initial phase of one-stream Transformer tracking approaches, tokens are generated from the patches of the target template and search regions, and then they are concatenated. Subsequently, attention among all these concatenated tokens is computed using the self-attention mechanisms of the encoder layers of the Transformer. Computing attention between all concatenated target template and search region tokens demonstrated promising performance by facilitating information flow \citep{ye2022joint} between template and search tokens, enabling both regions to extract target-specific features and accurately locate the target. As shown in Fig. \ref{fig:introduction}, previously proposed One-stream Transformer tracking approaches compute the attention between all target and search region tokens to enable the information flow and showed outstanding performances. However, no proper studies have been conducted to justify whether the attention of search region tokens on the target template tokens supports or reduces the discriminative ability of the trackers. Therefore, there is a need to investigate the attention mechanisms between different types of tokens in one-stream Transformer tracking.

Tracking a target object by only considering the initial target template is challenging, especially in long-term tracking scenarios where the target undergoes significant appearance changes. Some one-stream Transformer trackers \citep{cui2022mixformer, lan2022procontext} address this challenge by incorporating temporal cues of the target object through the use of dynamic templates in tracking. However, similar to other one-stream Transformer trackers, these trackers  concatenate the target template, dynamic template, and search region tokens, and then compute the attention between all concatenated tokens without investigating the impact of attention among different groups of tokens.

In this study, we investigate the information flow among the tokens of the target template, dynamic template, and search region in the one-stream Transformer tracking paradigm by systematically analyzing the attention between different groups of tokens. Based on experimental findings, it is evident that the information flow from a subset of search region tokens, which are generated from background patches, to the initial target template and dynamic target template tokens, reduces the tracker's discriminative capability. We have also observed that the information flow from the subset of search region tokens containing target cues to the initial and dynamic target template tokens enhances target-specific feature extraction and is also used to capture the appearance changes of the target. 

In previous tracking approaches, dynamic target templates are generated by tightly cropping the high-confidence target in an intermediate frame and then using those temporal cues to enhance tracking performance. In this work, we have also investigated how to maximize the utilization of temporal cues in the one-stream Transformer tracking paradigm. Our experimental study found that, instead of tightly cropping the dynamic target, cropping it with additional dynamic background cues enriches the tracking performances by providing valuable surrounding information of the target object. We have also found that including the dynamic background cues helps to accurately locate the target in the search region and is also used to avoid distractor objects in subsequent frames.

Based on our experimental analysis, we have proposed a novel one-stream Transformer tracker and called it as the Optimized Information Flow  Tracker (OIFTrack).  To utilize the temporal cues in tracking, the dynamic target template and dynamic background are included in the proposed OIFTrack approach. As the initial step of the tracker, tokens are generated from the patches of the initial target template, dynamic target template with the background region, and the search region and considered as four groups: initial target template tokens, dynamic target template tokens, dynamic background tokens, and search region tokens. In the early encoder layers of the proposed tracker, the information flow from the search region tokens and dynamic background tokens to the initial target template tokens, as well as the information flow from the search region tokens to the dynamic target tokens, is blocked to prevent disturbance to the tracker-specific feature extraction. In the deeper encoder layers of the proposed tracker, search region tokens are categorized into two groups: target  search tokens and non-target search regions, based on the attention scores of the search tokens with the initial and dynamic target template tokens. After the token grouping in deeper encoder layers, information flow from target search tokens to both initial and dynamic target template tokens is allowed to enrich the target-specific feature extraction and capture the target's appearance changes for more accurate localization. The proposed OIFTrack approach blocks the information flow between tokens using a simple attention masking technique. In summary, our contributions revolve around four fundamental aspects:
\begin{enumerate}  
	
	\item We propose an optimized information flow mechanism for the interaction among input tokens, achieved by dividing them into four groups in early layers and five groups in deeper layers, effectively blocking unnecessary information flows that disrupt the discriminative capability of the tracker.  
	
	\item We have incorporated dynamic background cues into the proposed tracker to avoid distractor objects and locate the target more accurately. To our knowledge, no previous Transformer trackers have considered the importance of dynamic background cues.
	
	\item In contrast to previous approaches employing learnable or complex modules, we have utilized a simple and efficient mechanism to partition the search tokens into target and non-target categories based on their attention scores from initial target and dynamic target tokens.

	\item We perform extensive experiments and evaluations to demonstrate the efficacy of our tracker. Our OIFTrack framework showed outstanding performance in various tracking benchmarks, such as GOT-10k \citep{huang2019got}, LaSOT \citep{Fan_2019_CVPR}, TrackingNet \citep{Muller_2018_ECCV}, and UAV123 \citep{UAV123_2016}.

\end{enumerate}

\section{Related Work}\label{sec2}
In this section, we review the literature on Transformer-based tracking approaches and information flow techniques related to the proposed optimized information flow mechanism.

\subsection{Two-stream Transformer Trackers}\label{sec2sec1}

Until recently, two-stream tracking pipeline-based approaches have shown state-of-the-art performances in VOT with high computational efficiency. In this pipeline of trackers, initial target template and search region features are independently extracted using a neural network model, and then the target is located in the search region by finding the similarity of target features in the search region features.

Transformers were initially introduced into the two-stream tracking pipeline as a feature fusion module \citep{chen2021transformer,gao2022aiatrack,song2022transformer,yan2021learning,yu2021high}, replacing the correlation modules of CNN-based Siamese trackers. In these CNN-Transformer-based two-stream trackers, a CNN backbone is employed to extract the features of both target template and search region. The extracted CNN features are then flattened and concatenated before being fed to a Transformer to capture feature dependencies using the attention mechanism. Finally, the enhanced search region features are input to a prediction head to locate the target. While some early CNN-Transformer trackers adopted Transformer architectures, such as DETR \citep{carion2020end} in STARK \citep{yan2021learning} tracker,  from the object detection task with minor modifications, recent approaches have recognized issues specific to Transformer-based tracking and subsequently adjusted their architectures. CNN-Transformer based trackers have demonstrated superior performance compared to CNN-based trackers by employing a trainable Transformer instead of the linear cross-correlation operation. 

Some of the recent two-stream Transformer trackers \citep{lin2022swintrack,fu2022sparsett} replaced the CNN backbone with two individual but identical Transformers to extract the features of both the target template and search region.  Since these trackers utilized the global feature representations of the Transformers in both feature extraction and future fusion they outperformed the CNN-Transformer based trackers in terms of accuracy. 

Although the two-stream Transformer trackers showed better performance than the CNN-based trackers due to the learnable feature fusion modules and the global feature learning capability of the Transformers, they extracted the target template and search region features individually. Therefore, the search region features are extracted without knowledge of the target cues, and as a result, the importance of target-specific features in the search region fails to be considered. To address this limitation, recent Transformer trackers enable information flow between target template and search region tokens from the early encoder layers, and these trackers are known as one-stream Transformer trackers.

\subsection{One-Stream Transformer Trackers}\label{sec2sec2}

One-stream Transformer trackers \citep{chen2022backbone, cui2022mixformer, ye2022joint,lan2022procontext, xie2023videotrack,gao2023generalized, Wu2023drop, Zhao2023representation, Wei2023autoregressive, Chen2023seqtrack} combine the feature extraction and future fusion processes and enable the bidirectional information flow between target template and search region tokens. In one-stream Transformer trackers, a set of encoder layers is utilized with the self-attention mechanism to jointly extract the features of the target and template tokens. Finally  , the features of the search region tokens from the last encoder layer are fed to a prediction head to locate the target. 
As these trackers enhance target-specific feature extraction, they improve the discriminative capability of the tracker, resulting in outstanding performances compared to both CNN-based trackers and two-stream Transformer based trackers.

In the last two years, a considerable number of one-stream Transformer trackers have been proposed, incorporating bidirectional information flow between target template and search region tokens. OSTrack \citep{ye2022joint} is well-known as one such tracker due to its outstanding performances in benchmark datasets and highly parallelized architecture.  It utilized a Masked Auto Encoder (MAE) \citep{He2022masked} based pre-trained model to initialize the encoder layers of the tracker, resulting in a remarkable performance boost. In addition, a candidate elimination module is included in OSTrack to remove background tokens from search regions, thereby increasing the computational efficiency of the tracker. Due to OSTrack's outstanding tracking performance in many benchmark datasets, several follow-up approaches \citep{lan2022procontext,xie2023videotrack,gao2023generalized} have been proposed within the one-stream tracking pipeline. Although OSTrack and follow-up approaches showed better performance than the two-stream Transformer trackers, they allowed bidirectional information flow between all target and search region tokens without evaluating the influence of tokens on discriminative target objects from the surroundings. As a result, their tracking performances are considerably limited. Moreover, since OSTrack and some of the follow-up approaches fail to consider temporal information, their tracking performances are considerably poor in long tracking sequences, especially when the target undergoes severe appearance changes.

A few recent approaches \citep{cui2022mixformer, lan2022procontext, xie2023videotrack, Chen2023seqtrack} have enhanced the tracking performances of one-stream Transformer trackers by incorporating the temporal cues of the target. In these trackers, dynamic target tokens are extracted from an intermediate frame with high-confidence detection and then concatenated with the initial target template and search region tokens. Some of these trackers follow special mechanisms to select the dynamic target template, such as the MixFormer \citep{cui2022mixformer} tracker, which used a learnable score prediction module to select reliable dynamic templates based on predicted confidence scores. Although these trackers enhance the tracking performance by incorporating dynamic tokens, they allow bidirectional information flow between all initial target, dynamic target, and search region tokens without considering their discriminative capability. In addition, most of these trackers fail to consider dynamic background cues, which would allow the tracker to incorporate surrounding information of the target for more accurate localization in the search frame.

In the proposed work, instead of allowing free bidirectional information flow between all target and search region tokens as in previous one-stream Transformer approaches, unnecessary information flow between tokens is blocked to enhance the discriminative capability of the tracker. Additionally, surrounding information of the dynamic target is included in the proposed work to accurately locate the target and avoid distractor objects in the search frame. 

\subsection{Information Flow Variants in One-Stream Tracking}\label{sec2sec3}

All of the one-stream Transformer trackers freely allowed bidirectional information flow between target and search region tokens in encoder layers without any restrictions. However, recently, a few trackers have controlled the information flow between tokens for various purposes.  The recently proposed SeqTrack \citep{Chen2023seqtrack} and ARTrack \citep{Wei2023autoregressive} approaches restrict the information flow between tokens in decoder layers to prevent tokens from attending to subsequent tokens. However, in the encoder layer of these trackers, attention features are extracted by freely allowing bidirectional information flow between target and search region tokens.

The GRM tracker \citep{gao2023generalized} restricts the information flow between the template and search tokens by allowing only a subset of search region tokens, determined by the selection of a learnable adaptive token division module, to interact with template tokens.  The GRM approach performs as a two-stream tracker in some encoder layers, as all search tokens are rejected by the division module for interaction with template tokens in those layers, and performs as a one-stream tracker in some other layers, as all search tokens are selected for interaction. While functioning as a two-stream tracker in certain encoder layers, the GRM tracker individually extracts attention features from the target and search region tokens, which may result in the failure to extract target cues from the search tokens in those layers, potentially leading to a decrease in tracking performance.

Recently proposed, the F-BDMTrack approach \citep{Yang_2023_ICCV} blocks the information flow between the foreground and background tokens of the target template using a fore-background agent learning module. Additionally, the information flow between the foreground and background tokens of the search region is also restricted using that module to enhance the discriminative capability of the tracker. After the fore-background module, a distribution-aware attention module is used to prevent the incorrect interaction between the foreground and background tokens by modeling foreground-background distribution similarities. In this tracker, ground-truth bounding boxes are used to identify the foreground and background tokens of the template, while a pseudo bounding box generation technique is employed to identify the foreground in the search region.

Although the GRM and F-BDMTrack approaches block information flow between different groups of tokens for various purposes, they employ a learnable module for token grouping from early encoder layers. Therefore, a search token may be identified as a potential target token in one layer and as a background token in another, due to variations in the knowledge level of the learnable module between early and deeper layers, that may lead to poor tracking performance. In contrast to previous approaches, the proposed OIFTrack approach is carefully designed with a simple and neat architecture. Additionally, by conducting the search token partitioning only in deeper encoder layers using the learned knowledge from the middle encoder layers and blocking the information flow accordingly, it achieves more accurate identification of target and non-target tokens from the search region in a simple and fast manner.

\section{Methodology}\label{sec3}
In this section, we introduce our one-stream Transformer tracking approach, termed as OIFTrack, which relies on optimizing information flow between tokens. Initially, we delve into the specifics of the baseline tracking model, followed by an explanation of how the proposed tracker utilizes the temporal cues for enhanced tracking. Subsequently, we provide a detailed description of the proposed optimized information flow mechanism. Finally, we outline the background token elimination technique and details of the prediction head within this section.

\subsection{Baseline Tracking Model}\label{subsec1}
In this section, we explore the baseline tracking model used in one-stream Transformer trackers. All of the previous approaches \citep{ye2022joint, chen2022backbone, cui2022mixformer} employed this baseline model to establish their tracking methodologies.   

In the one-stream Transformer tracking paradigm, a tracker takes the initial target template $Z \in \mathbb{R}^{H_Z \times W_Z \times 3}$ and search region  $X \in \mathbb{R}^{H_X \times W_X \times 3}$ as inputs, and then produces the location of the target in the search region as the output. As the first step, template and search region images are divided into non-overlapping patches, each of size $P \times P$.  The total number of target template patches ($N_Z$) and search region patches ($N_X$) can be calculated as follows:
\begin{equation}
	\begin{split}
		N_Z = \frac{H_Z \times W_Z }{P^2} \\
		N_X = \frac{H_X \times W_X }{P^2}
	\end{split}
\end{equation}

After partitioning the patches, a learnable linear projection layer is employed to generate template and search patch embeddings from their respective patches. Subsequently, the spatial information of the template and search patches is individually embedded using a learnable position embedding scheme. The outcome of the embeddings is known as tokens, and we refer to the target template and search region tokens as $E_Z$ and $E_X$, respectively. Before being fed to the encoder layers of the Transformer, these tokens are concatenated to generate a sequence of tokens [$E_Z$ ; $E_X$] with a total length of $N_Z + N_X$.

In the next phase of the one-stream Transformer tracking pipeline, attention features for the concatenated tokens are computed through a multi-head attention (MHA) block and a then fed to a feed-forward network (FFN) in each encoder layer.  The self-attention of $i^{th}$ head in an encoder layer is computed as follows:
\begin{equation}\label{eq2}
	\mathrm{Attention}(Q_i,K_i,V_i)=\mathrm{Softmax}\left(\frac{Q_i\cdot K_i^T}{\sqrt{d_{k_i}}}\right)\cdot V_i
\end{equation}
where, Query ($Q_i$), Key ($K_i$) and Value ($V_i$) are the three linear projections of the input matrix and  $d_{k_i}$ represents the feature dimension of $K_i$. The MHA block of the one-stream tracker captures a wide range of relationships and dependencies between input tokens through linear projection while concurrently processing them with multiple independent attention heads. The process of MHA can be formulated as follows:

\begin{equation}
	\begin{split}
		& \mathrm{MHA}(Q,K,V) = \mathrm{Concat}(\mathrm{head}_{1} , ...,\mathrm{head}_{h})\cdot W_O \\
		& \mathrm{head}_{i} = \mathrm{Attention}(Q_i,K_i,V_i)
	\end{split}
\end{equation}
where $h$ is the number of attention heads in the tracker and  $W_O$ denotes the output projection matrix.

In the $n^{th}$ encoder layer of the baseline tracking model, the concatenated target and search region tokens ($[E_Z^n;E_X^n]$)  are obtained from previous layer and then considered as the input tokens. After a normalization operation, the attention features of these tokens are computed by the MAH block and then added to the input tokens. Following another normalization operation, the combined features are fed into the FFN block of the encoder layer to obtain the output tokens. In summary, the operations of the $n^{th}$ encoder layer can be expressed as:

\begin{equation}
	\label{eqn:attentaion}
	\begin{split}
		&  [E_Z^{'n} ; E_X^{'n}] = [E_Z^n ; E_X^n] + \mathrm{MHA}(Q,K,V) \\
		& [E_Z^{n+1} ; E_X^{n+1}]= [E_Z^{'n}  ;E_X^{'n}] + \mathrm{FFN}([E_Z^{'n} ; E_X^{'n}])
	\end{split}
\end{equation}
where $[E_Z^{n+1} ; E_X^{n+1}]$ is the output tokens of the $n^{th}$ encoder layer. 

From the output tokens ($[E_Z^{l+1}; E_X^{l+1}]$) of the last encoder layer of the baseline tracking model, the search region tokens ($[E_X^{l+1}]$) are extracted and fed into a prediction head to locate the target within the search region.

Most one-stream Transformer tracking approaches \citep{ye2022joint, chen2022backbone, Wu2023drop, Zhao_2023_CVPR} only utilize the baseline tracking model with free bidirectional information flow. In contrast, the proposed OIFTrack approach goes further by enhancing the tracking performance of the baseline tracker through the incorporation of temporal cues and the optimization of information flow between the tokens.

\subsection{Temporal Cues Utilization}\label{subsec2}

Capturing and utilizing the temporal information is important in VOT, especially in long-term tracking scenarios where the appearance of the target may undergo significant changes. While many one-stream Transformer trackers \citep{ye2022joint, chen2022backbone, Wu2023drop, Zhao_2023_CVPR} have shown better performance on benchmark datasets, their effectiveness is considerably limited by their failure to incorporate temporal cues in tracking, relying solely on the initial target template to locate the target in long-term sequences. Although some recent approaches \citep{lan2022procontext, cui2022mixformer, xie2023videotrack} have included temporal cues by incorporating dynamic templates, our focus is on further enriching the utilization of temporal cues within the one-stream Transformer tracking paradigm.

Background cues are important in VOT as they are used to capture the surrounding information of the target object. Some Transformer tracking approaches \citep{mayer2022transforming, wang2021transformer, He_Zhang_Xie_Li_Wang_2023} expand the size of the target template to match the size of the search region, even though it increases the computational complexity of the tracker, aiming to incorporate the surrounding background information of the target. However, it is observed that the surrounding information of the initial target template becomes less useful for locating the target after a set of frames, especially in long-term tracking scenarios. Moreover, these approaches extract features from both the actual target region and a large portion of background regions together in the template, resulting in the distraction of target-specific features. Therefore, in the proposed approach, we have utilized the background cues from the dynamic template since they are very close to the surrounding region of the target in the search region while keeping the initial template without any additional background patches. 

\begin{figure}[t]%
	\centering
	\includegraphics[width=0.48\textwidth]{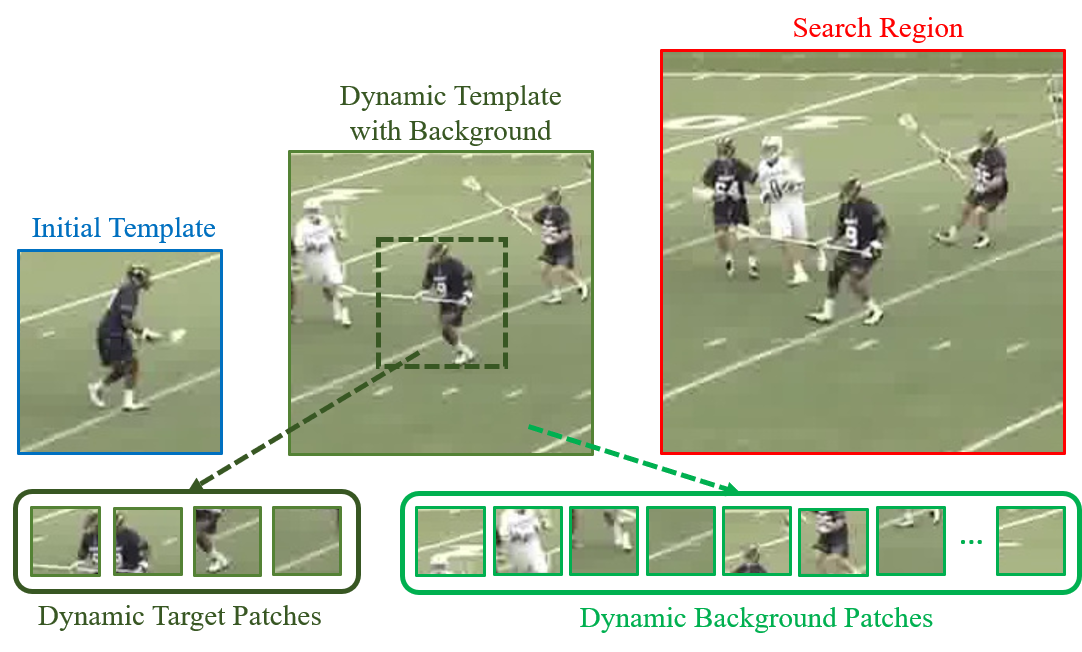}
	\caption{Illustration of the proposed temporal cue utilization mechanism. We have extracted the temporal cues of the target as dynamic target patches and the surrounding cues as dynamic background patches. Since similar objects from the surrounding region are also identified as dynamic background patches, the proposed tracker can easily avoid distractor objects and accurately locate the target in the search region. }
	\label{fig:dynamic}
\end{figure}

In the proposed OIFTrack approach, we include the dynamic target template from an intermediate frame in which the target object was located with high confidence. In addition to the dynamic target template, as shown in Fig. \ref{fig:dynamic}, we extract a few additional patches from the surrounding region of the dynamic target and consider them as dynamic background patches. Since nearby similar objects are also identified as dynamic background patches, the proposed tracker is able to successfully avoid distractor objects in the search region. In addition, unlike previous approaches, the proposed tracker restricts the information flow from dynamic background patches to the initial target template patches, ensuring that the target-specific features of the initial template remain undistracted.

In the proposed tracker, similar to the initial target template and search region patches, the dynamic target template image with background is divided into non-overlapping patches, each of size $P \times P$. After the linear projection and positional embedding, dynamic target template tokens and dynamic background tokens are generated; we refer to them as $E_{DT}$ and $E_{DB}$, respectively.  Before being fed to the encoder layers of the Transformer, these tokens are concatenated with the initial target template and search region tokens to generate a sequence of input tokens [$E_Z$; $E_{DT}$; $E_{DB}$; $E_X$]. In the first frame of a tracking sequence, the dynamic target and dynamic background tokens are initialized by incorporating the target and surrounding cues in the initial frame. In every fixed frame interval, based on the highest confidence score, these tokens are updated to obtain the temporal cues.

\subsection{Optimized Information Flow between Tokens}\label{subsec3}

\begin{figure*}[t]%
	\centering
	\includegraphics[width=\textwidth]{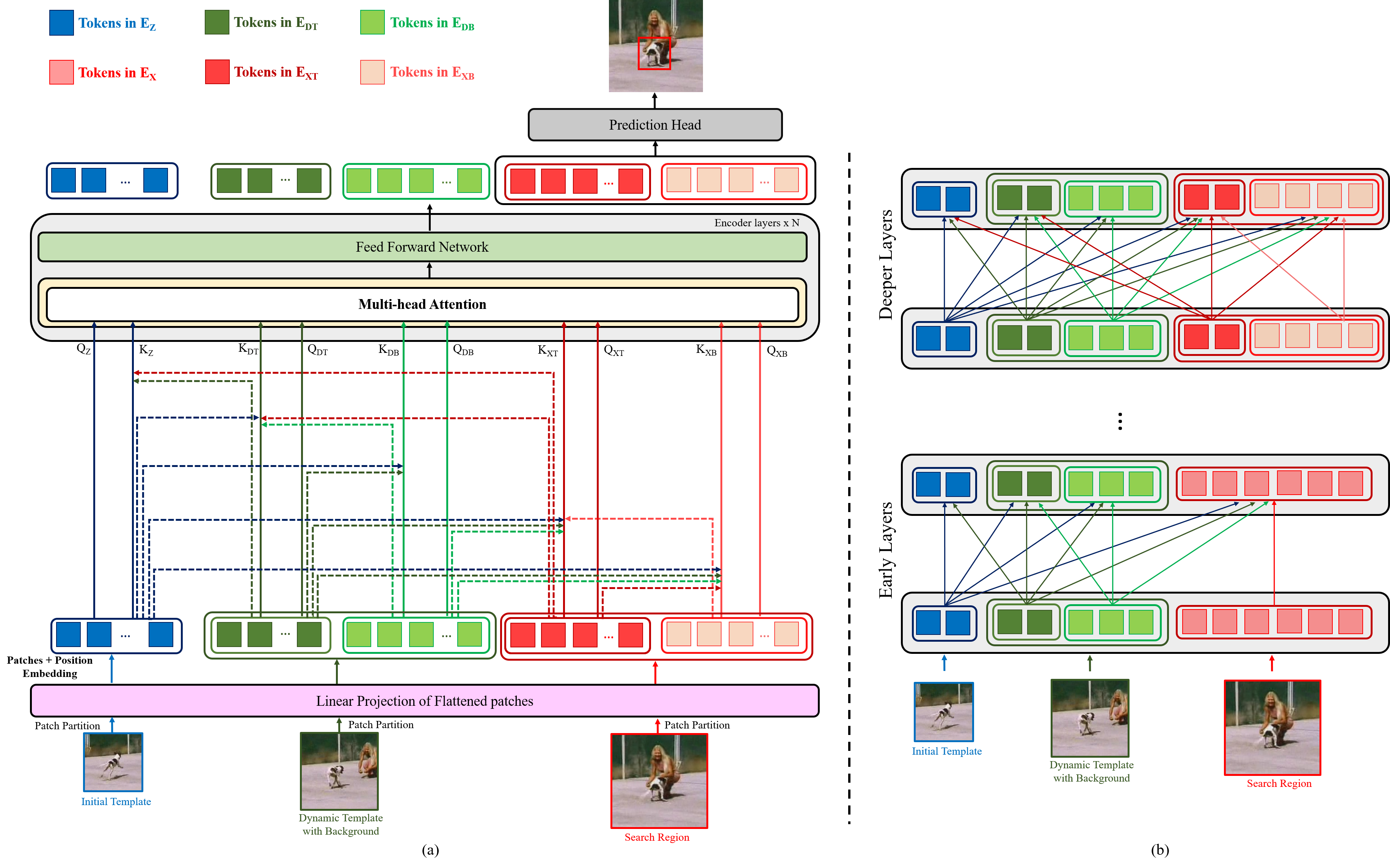}
	\caption{(a): The overall tracking framework of the proposed tracker. The diagram illustrates the proposed attention feature extraction mechanism only in the last few encoder layers, and values are omitted for clarity. (b): Illustration of the optimized information flow between tokens in early encoder layers and deeper encoder layers.}
	\label{fig:methodology}
\end{figure*}

We have conducted a comprehensive analysis of the information flow between target template and search tokens, focusing on enhancing the performance of one-stream tracking further, and have discovered several informative findings. In the search region tokens, the number of non-target tokens is higher compared to those containing the target cues, as all of these trackers search for the target in a large region. In many frames, most of the non-target search tokens contain background cues, while a few others contain cues of distractor objects. We have found that the information flow from the large number of background search tokens to the fewer number of target template tokens gives more influence to the background rather than the target, thereby reducing the importance of target-specific features in the template. In addition, information flow from search tokens with distractor cues reduces the originality of target-specific cues in the template, leading to drifting problems. However, we have also found that bidirectional interaction between target template tokens and search tokens containing target cues is crucial, as it is used to enrich the target-specific feature extraction, enabling the tracker to capture the appearance changes of the target.

We have expanded the comprehensive analysis of information flow between tokens by incorporating dynamic target tokens and dynamic background tokens. Consistent with previous findings, the information flow from dynamic background tokens to the initial target template tokens reduces the importance of target-specific features, while bidirectional information flow between dynamic target template and initial template tokens strengthens target-specific feature extraction.

Based on the findings, the proposed OIFTrack approach, instead of employing the free bidirectional information flow used in previous approaches, optimizes the information flow by blocking certain unnecessary token interactions that might distract the discriminative capability of the tracker. In the early encoder layer of the OIFTrack approach, the concatenated tokens are categorized as initial target template tokens, dynamic target template tokens, dynamic background tokens, and search tokens, and then the unnecessary information flows between these categories are blocked. The search tokens are maintained as a single category in the early layers because these layers lack sufficient knowledge about target-specific cues to perform partitioning. As shown in Fig. \ref{fig:methodology}, in the deeper encoder layer of the proposed work, the number of token categories is increased by partitioning the search region tokens into non-target tokens and tokens with target cues, and then unnecessary information flows between these categories are blocked.   We blocked the unnecessary information flow between tokens by utilizing the attention masking technique. In this technique, some attention scores in the matrix are manually set to negative infinity to prevent the interaction between corresponding tokens. The attention score matrix ($S$) is computed as follows:

\begin{equation}
	S = Q \cdot K^T 
\end{equation}
since the Softmax operation is used to calculate the attention features by using the score matrix, as given in Equation \ref{eq2}, the information flow between the selected tokens is completely ignored in the proposed tracker.

The proposed OIFTrack approach obtained the initial target ([$E_Z$]), dynamic target ([$E_{DT}$]), dynamic background ([$E_{DB}$]), and search region ($E_X$) tokens and then concatenated them as a sequence of input tokens ([$E_Z$; $E_{DT}$; $E_{DB}$; $E_X$]). In an early encoder layer $n$, the attention features of the initial target template tokens are computed by obtaining the Query ($Q_Z$) from the initial template tokens and the Key ($K_Z$) and Value ($V_Z$) from both the initial target template and dynamic target template tokens, while ignoring the information flow from both the search tokens and dynamic background tokens. The attention feature extraction process of initial target template tokens in the $n^{th}$ layer can be expressed as follows: 

\begin{equation}
	\label{eqn:z}
	\begin{split}
		&	Q_Z  \leftarrow E_Z^{n} \\
		&	K_Z ,  V_Z  \leftarrow [E_Z^{n};E_{DT}^{n}] \\
		& E_Z^{'{n}}  = E_Z^{n} + MHA(Q_Z,K_Z,V_Z)		
	\end{split}
\end{equation}

In that $n^{th}$ early encoder layer, attention features of dynamic tokens is computed by obtaining the Queries ($Q_{DT}$ and $Q_{DB}$) from corresponding dynamic tokens and the Keys ($K_{DT}$ and $K_{DB}$) and Values ($V_{DT}$ and $V_{DB}$) from initial target template, dynamic target template, and dynamic background tokens, while ignoring the information flow from the search tokens. The attention feature extraction process of dynamic template and background tokens in an early layer can be expressed as follows:

\begin{equation}
	\label{eqn:dt}
	\begin{split}
		&	Q_{DT}  \leftarrow E_{DT}^{n} \\
		&	Q_{DB}  \leftarrow E_{DB}^{n} \\
		&	K_{DT} , V_{DT} \leftarrow [E_Z^{n};E_{DT}^{n};E_{DB}^{n}] \\
		&	K_{DB} , V_{DB}  \leftarrow [E_Z^{n};E_{DT}^{n};E_{DB}^{n}] \\
		& E_{DT}^{'{n}} = E_{DT}^{n} + MHA(Q_{DT},K_{DT} , V_{DT})		\\
		& E_{DB}^{'{n}} = E_{DB}^{n} + MHA(Q_{DB},K_{DB} , V_{DB})
	\end{split}
\end{equation}

In that $n^{th}$ early encoder layer, attention feature extraction process of search tokens can be expressed as follows:

\begin{equation}
	\label{eqn:x}
	\begin{split}
		&	Q_X  \leftarrow E_X^{n} \\
		&	K_X , V_X  \leftarrow [E_Z^{n};E_{DT}^{n};E_{DB}^{n}; E_X] \\
		& E_X^{'{n}} = E_X^{n} + MHA(Q_X, K_X , V_X)		
	\end{split}
\end{equation}

As the last step of the feature extraction in the $n^{th}$ early encoder layer, the extracted attention features are concatenated and then fed to the FFN block. This process can be expressed as:

\begin{equation}
	\label{eqn:all}
	\begin{split}
		& [E_Z^{{n}+1};E_{DT}^{{n}+1};E_{DB}^{{n}+1};E_X^{{n}+1}]  = \\ &[E_Z^{'{n}};E_{DT}^{'{n}};E_{DB}^{'{n}};E_X^{'{n}}] + FFN([E_Z^{'{n}};E_{DT}^{'{n}};E_{DB}^{'{n}};E_X^{'{n}}])		
	\end{split}
\end{equation}

where $[E_Z^{{n}+1};E_{DT}^{{n}+1};E_{DB}^{{n}+1};E_X^{{n}+1}]$ are the output tokens of the $n^{th}$ encoder layer. 

In the deeper encoder layers of the proposed approach, search region tokens are partitioned into target search tokens and non-target search tokens. This partition is carried out based on the attention scores of the search region tokens in relation to the initial target template and dynamic target template tokens. As the initial step in this partitioning process, attention score weights of all search region tokens are computed as follows:
\begin{equation}
	\label{eqn:token}
	\omega = 	\phi(\mathrm{Softmax}\left(\frac{Q_Z\cdot K_X^T}{\sqrt{d_\mathbf{k}}}\right)+
	\mathrm{Softmax}\left(\frac{Q_{DT}\cdot K_X^T}{\sqrt{d_\mathbf{k}}}\right)) \in \mathbb{R}^{1 \times N_X}
\end{equation}
We have only used the center region tokens of the initial target and dynamic target in the above computation to accurately identify the target  search tokens. Since the obtained weighted score matrix ($\omega$) represents the relevance of search region tokens to the initial target and dynamic target template tokens, we sorted the scores and selected the corresponding top-K tokens as search tokens with target cues, while the remaining were considered as non-target search tokens.

After the search token partitioning, a deeper encoder layer $m$ receives the initial target ([$E_Z$]), dynamic target ([$E_{DT}$]), dynamic background ([$E_{DB}$]), target search ([$E_{XT}$]), and non-target search ([$E_{XB}$]) tokens from previous layer and then concatenated them as a sequence of tokens ([$E_Z$; $E_{DT}$; $E_{DB}$; $E_{XT}$; $E_{XB}$]). The attention features of the initial target tokens are extracted by blocking the information flow from both the dynamic background tokens and the non-target search tokens, as described in the following equation:

\begin{equation}
	\label{eqn:z_d}
	\begin{split}
		&	Q_Z \leftarrow E_Z^{m} \\
		&	K_Z, V_Z  \leftarrow [E_Z^{m};E_{DT}^{m};E_{XT}^{m}] \\
		& E_Z^{'{m}} = E_Z^{m} + MHA(Q_Z,K_Z, V_Z)		
	\end{split}
\end{equation} 
In that deeper encoder layer, the attention features of the dynamic target tokens are computed by blocking the information flow from non-target search tokens, and this process is expressed as follows:
\begin{equation}
	\label{eqn:dt_d}
	\begin{split}
		&	Q_{DT}  \leftarrow E_{DT}^{m} \\
		&	K_{DT}  , V_{DT}   \leftarrow [E_Z^{m};E_{DT}^{m};E_{DB}^{m};E_{XT}^{m}] \\
		& E_{DT}^{'{m}} = E_{DT}^{m} + MHA(Q_{DT},K_{DT},V_{DT})		
	\end{split}
\end{equation}
Additionally, the attention features of the dynamic background tokens are extracted by restricting the information flow from all search region tokens, as follows:
\begin{equation}
	\label{eqn:db}
	\begin{split}
		&	Q_{DB}  \leftarrow E_{DB}^{m} \\
		&	K_{DB} , V_{DB}  \leftarrow [E_Z^{m};E_{DT}^{m};E_{DB}^{m}] \\
		& E_{DB}^{'{m}} = E_{DB}^{m} + MHA(Q_{DB},K_{DB},V_{DB})		
	\end{split}
\end{equation}
The attention features of the search tokens ($E_X = [E_{XT};E_{XB}]$) are obtained by allowing interaction with all other tokens, and this process can be written as: 
\begin{equation}
	\label{eqn:x_d}
	\begin{split}
		&	Q_X  \leftarrow [E_{XT}^{m};E_{XB}^{m}] \\
		&	K_X , V_X  \leftarrow [E_Z^{m};E_{DT}^{m};E_{DB}^{m}; E_{XT}^{m};E_{XB}^{m}] \\
		& [E_{XT}^{'{m}};E_{XB}^{'{m}}] = [E_{XT}^{m};E_{XB}^{m}] + MHA(Q_X,K_X,V_X)		
	\end{split}
\end{equation}
As the last step of the feature extraction in the $m^{th}$ deeper encoder layer, the extracted attention features are concatenated and then fed to the FFN block. This process can be expressed as:

\begin{equation}
	\label{eqn:all_d}
	\begin{split}
		& [E_Z^{{m}+1};E_{DT}^{{m}+1};E_{DB}^{{m}+1};E_{XT}^{{m}+1},E_{XB}^{{m}+1}]  = \\ &[E_Z^{'{m}};E_{DT}^{'{m}};E_{DB}^{'{m}};E_{XT}^{'{m}};E_{XB}^{'{m}}] +\\ &FFN([E_Z^{'{m}};E_{DT}^{'{m}};E_{DB}^{'{m}};E_{XT}^{'{m}};E_{XB}^{'{m}}])		
	\end{split}
\end{equation}

Finally, from the output tokens of the last encoder layer of the proposed tracker, search region tokens are separated and then fed to a prediction head to locate the target. The overall tracking framework of the proposed OIFTrack approach is depicted in Fig. \ref{fig:methodology}. It illustrates the attention feature extraction mechanism of the deeper encoder layers on the left side while demonstrating the difference in information flow between early and deeper encoder layers on the right side. Since the proposed optimized information flow mechanism enables bidirectional information flow among all tokens extracted from the initial target, dynamic target, and target search regions, the proposed tracker can effectively capture the appearance changes of the target object compared to previous approaches. Moreover, as the proposed optimized information flow mechanism blocks information flow from tokens containing non-target cues to tokens containing target cues, target-specific features are extracted more accurately, thereby enriching the discriminative capability of the tracker.

\subsection{Background Token Elimination}\label{subsec4}
While the proposed OIFTrack approach enhances tracking accuracy, its computational efficiency is reduced due to the increase in the total number of input tokens resulting from the incorporation of tokens from the dynamic template. To address this issue, we eliminate a set of non-target search tokens in some encoder layer of the proposed tracker. 

In one-stream Transformer tracking, a large portion of input tokens comes from the search region, with many of them containing only background cues. Therefore, eliminating the background tokens from the concatenated input tokens in deeper layers is one of the strategies to reduce the number of computations. Although the background token elimination mechanism is utilized in some previous approaches \citep{ye2022joint,lan2022procontext}, we incorporated dynamic target cues to accurately identify the search tokens containing only background cues. Similar to the search token partitioning, Equation \ref{eqn:token} is used to identify the search region tokens that have background cues, based on their lowest attention scores relative to the initial and dynamic template tokens. Then, among the tokens with the lowest scores, $P$ number of tokens are selected based on their poor attention scores and subsequently eliminated from the input token set.

\subsection{Prediction Head}\label{subsec5}

The prediction head of the proposed tracker obtains the search region features from the final encoder layer and then reshapes them back into two dimensions with the size of $(W_x \times H_x)$ to train the convolution-based heads. Similar to other single-stream Transformer trackers \citep{ye2022joint,lan2022procontext}, the prediction head of the proposed OIFTrack consists of three components: a classification head, an offset head, and a size head. These heads are used to generate classification scores, correct discretization errors, and predict the width and height of the target, respectively. The target state in the search region is determined by selecting the position with the highest confidence classification score and incorporating the corresponding offset and size coordinates to calculate the size of the bounding box at that position.

The classification and regression losses of the prediction head components are used to train the proposed tracker. The overall loss function is calculated as a weighted combination of the losses of components: a focal loss \citep{law2018cornernet} for the classification head ($L_{cls}$), a generalized IoU loss ($L_{giou}$) \citep{rezatofighi2019generalized} for the offset head, and an $L_1$ loss for the size head. It can be expressed as: 

\begin{equation}
	\label{eqn:loss}
	Loss = 	L_{cls} +\lambda_{iou}L_{iou} + \lambda_{L_1}L_{1}
\end{equation}

where $\lambda_{iou}$ = 2 and $\lambda_{L_1}$ = 5 are the regularization parameters in our experiments as in \citep{yan2021learning}.

\section{Experiments}\label{sec4}

\subsection{Implementation Details}\label{sec3subsec1}
The proposed OIFTrack approach is implemented in Python using the PyTorch framework, and both training and evaluation are conducted on a Tesla P100 GPU with 16 GB of memory.

The architecture of the proposed tracker, except for the prediction head, is similar to the encoder part of ViT-B \citep{dosovitskiy2021image}. Similar to many other one-stream trackers, we utilize the pre-trained MAE \citep{He2022masked} model as the backbone to initialize the Transformer architecture of the proposed approach. In our tracker, the search and target template sizes are set as $256 \times 256$ pixels and $128 \times 128$ pixels, respectively. Both are divided into non-overlapping patches, each with a size of $16 \times 16$ pixels. Although some one-stream trackers trained their model in two variants with search region sizes of $256 \times 256$ and $384 \times 384$ pixels, we did not train the other variant due to limitations in our GPU memory. In an intermediate frame of the proposed tracker, the dynamic region is cropped, resized to $192 \times 192$ pixels, with the central $128 \times 128$ pixels assigned for the dynamic target, and the remaining surrounding pixels assigned for the dynamic background. In addition, in the proposed tracker, the  token partitioning for the search tokens was conducted from the $10^{th}$ encoder layer.

The proposed OIFTrack approach was trained on the training sets of LaSOT \citep{Fan_2019_CVPR}, TrackingNet \citep{Muller_2018_ECCV}, GOT-10k \citep{huang2019got} and COCO 2017 \citep{coco_2014} benchmarks to evaluate the performance on the testset of LaSOT, UAV123 and TrackingNet. In addition, our approach was exclusively trained on the GOT-10k training set to evaluate its performance on the test set of that dataset, following the protocol described in \citep{huang2019got}. During training, we employ a few data augmentation techniques such as brightness jittering and horizontal flipping to enhance the generality of the tracker. The AdamW \citep{loshchilov2018decoupled} optimizer is used to train the model. The proposed tracker is trained for 300 epochs, with 60k image pairs used in each epoch. The model was trained with a learning rate of 1e-4, and it was decreased by a factor of 10 after 240 epochs. For the GOT-10k dataset, training was conducted for 100 epochs, and the learning rate was decreased by a factor of 10 after 80 epochs.

\subsection{Evaluation Protocols}\label{sec3subeva}
Similar to other VOT approaches, we have utilized the percentage of average overlap (AO), Success Rate at a 0.5 threshold (SR$_{0.5}$), and the Success Rate at a 0.75 threshold (SR$_{0.75}$) to measure tracking performances on the GOT-10k dataset. For other benchmarks, we use the percentage of Area Under Curve (AUC) score, precision (P), and normalized precision (P$_n$) to measure performances.

\subsection{Ablation Studies}\label{sec3subsec2}

\begin{figure}[t]%
	\centering
	\includegraphics[width=0.48\textwidth]{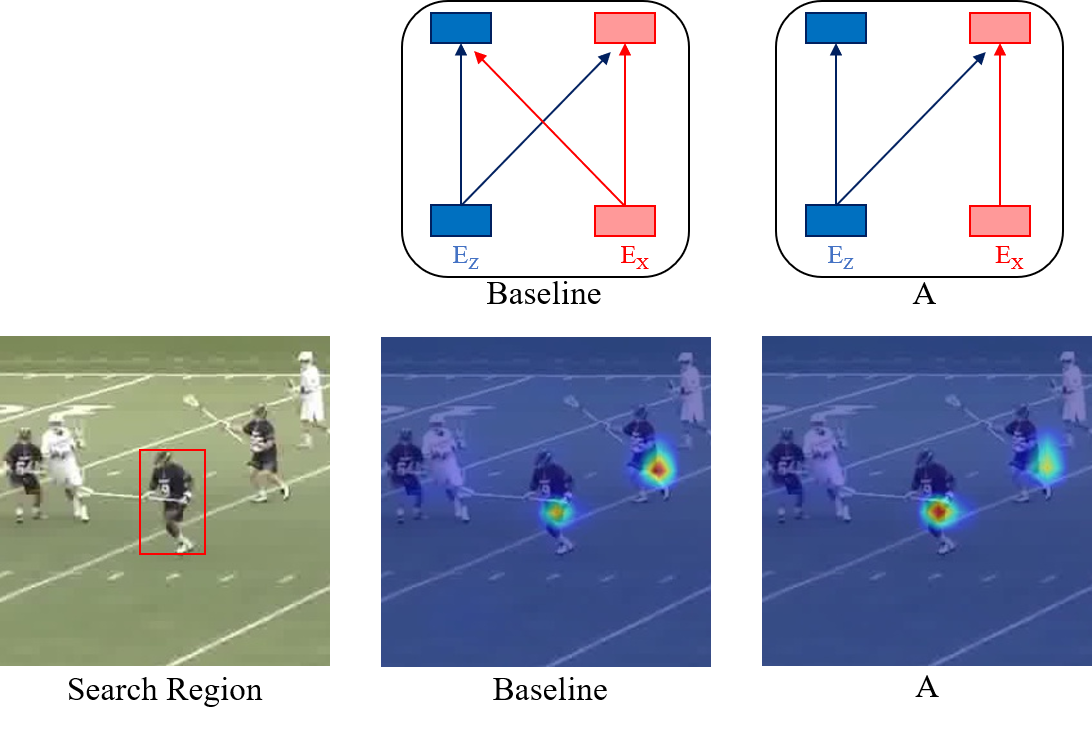}
	\caption{Top: Information flow diagram for the baseline model and model A, which blocks the interaction from search tokens ($E_{X}$) to target template tokens ($E_{Z}$). Bottom: Corresponding heatmap illustration of the search region for a tracking sequence. }
	\label{fig:Ablation1}
\end{figure}

\begin{table}[t]
	\footnotesize
	\begin{center}
		\caption{Study on blocking the information flow from search tokens ($E_{X}$) to target template tokens ($E_{Z}$). }\label{tabel:Ablation1}%
		\begin{tabular}{cclll}
			\toprule
			\multirow{2}{*}{\textbf{Model}}& \multirow{2}{*}{\textbf{Information flow}} & \multicolumn{3}{c}{\textbf{GOT-10k}}\\ 
			\cmidrule{3-5}
			& &  AO & $\mathrm{SR_{0.5}}$ & $\mathrm{SR_{0.75}}$ \\
			\midrule
			Baseline & Free Bidirectional & 71.0 & 80.1 & 67.3 \\
			\midrule
			A & $E_{X}$ to $E_{Z}$ blocked  & \textbf{71.4} &	\textbf{80.6} &	\textbf{67.8} \\
			
			\midrule
		\end{tabular}
	\end{center}
\end{table}

\subsubsection{Blocking Information flow from Search to Target Template Tokens}\label{sec3subsec2subsec1}

We have conducted detailed ablation studies to justify the design of the proposed tracker and to verify the choice of parameters. The ablation experiments were conducted by training the models on the training set of the GoT-10K dataset and then comparing their performances on its test set. Additionally, we utilized the  heatmap of the classification head to demonstrate the effectiveness of the proposed tracker.

Based on our comprehensive analysis of token interactions, we found that allowing bidirectional information flow from search region tokens to target template tokens reduces the discriminative capability of the tracker. To justify these findings, we trained a model, denoted as model A in Fig. \ref{fig:Ablation1}, by blocking the information flow from search tokens to target template tokens and compared its performance with that of the baseline model, which freely allows bidirectional information flow between all tokens.

Based on the experimental results in Table \ref{tabel:Ablation1}, since the information flow from a large number of non-target tokens in the search region reduces the importance of target-specific features, the performance of the baseline model is poorer than that of model A. In addition, the visualization of the heatmap in Fig. \ref{fig:Ablation1} also justifies the blocking by showing that the classification scores of the distractor object are higher in the baseline model compared to model A. 

\subsubsection{Importance of the Dynamic Background Cues}\label{sec3subsec2subsdy}

Differing from other approaches, we have utilized dynamic target and dynamic background cues to capture the temporal appearance changes of the target and learn the surrounding information of the target, respectively. From an adjacent high-confidence detection frame, dynamic target and background cues are extracted in the proposed tracker and used to incorporate temporal information. To justify our design concept, we trained two separate models, referred to as model B and model C, by including dynamic template cues and both dynamic target and background cues, respectively. 

\begin{figure}[t]%
	\centering
	\includegraphics[width=0.48\textwidth]{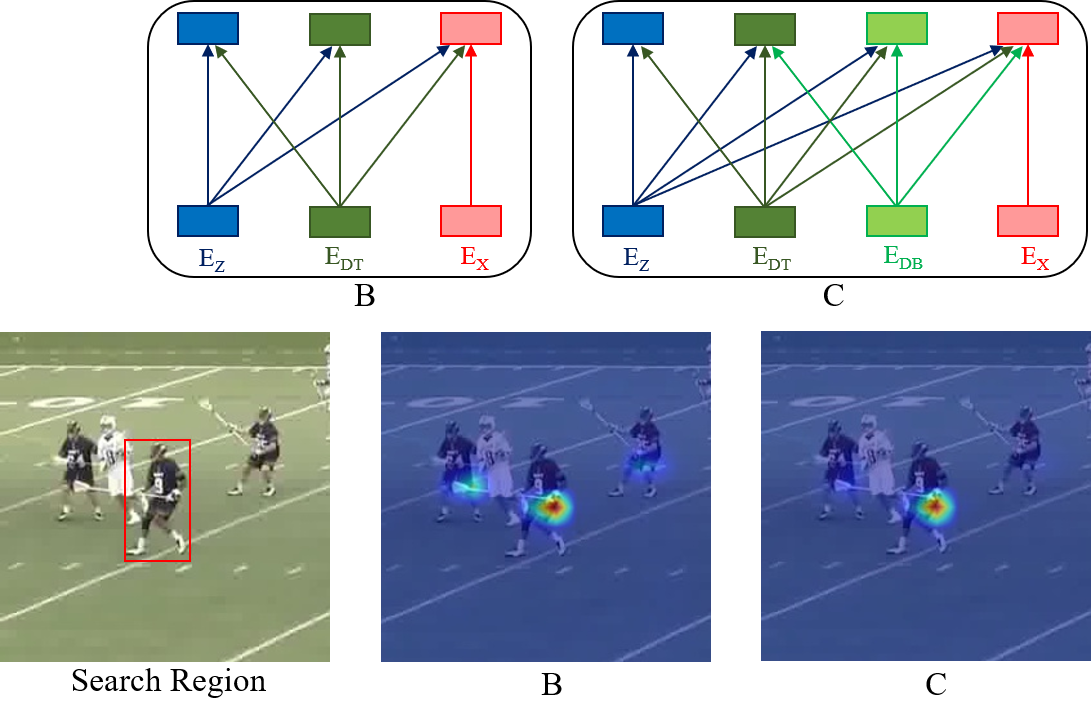}
	\caption{Top: Information flow diagrams for Model B, which includes dynamic target ($E_{DT}$) cues, and for Model C, which includes both dynamic target and background ($E_{DB}$) cues. Bottom: Corresponding heatmap visualization of the search region based on the classification scores. }
	\label{fig:Ablation2}
\end{figure}

Experimental results in Table \ref{tabel:Ablation2} clearly show that including the dynamic target  cues enhances tracking performance, as evidenced by the comparison between Model A and Model B. However, the substantial increase in tracking performance from Model B to Model C clearly indicates the importance of dynamic target cues in tracking. In addition, to justify the inclusion of dynamic background cues in the proposed tracker, we have compared the heatmap of classification scores for a search region in Model B and Model C. As shown in Fig. \ref{fig:Ablation2}, while Model B equally assigns importance to a nearby distractor object, Model C effectively ignores that distractor by capturing the surrounding information of the target from dynamic background cues. 

\begin{table}[!t]
	\footnotesize
	\begin{center}
		\caption{Study on the utilization of temporal cues. The involvement of the initial target ($E_Z$), dynamic target ($E_{DT}$), dynamic background ($E_{DB}$), and search tokens ($E_{X}$) in each tracking model, along with their performances, is provided.}\label{tabel:Ablation2}%
		\begin{tabular}{cM{0.1cm}M{0.2cm}M{0.2cm}M{0.1cm}M{1.7cm}M{0.35cm}M{0.4cm}M{0.5cm}}
			\toprule
			\multirow{2}{*}{\textbf{Model}}& \multicolumn{4}{c}{\textbf{Token Groups}} & \multirow{2}{*}{\textbf{Blocked}}& \multicolumn{3}{c}{\textbf{GOT-10k}}\\ 
			\cmidrule{2-5} \cmidrule{7-9}
			& $E_Z$ & $E_{DT}$ & $E_{DB}$ & $E_X$ & \textbf{Information Flow} & AO & $\mathrm{SR_{0.5}}$ & $\mathrm{SR_{0.75}}$ \\
			\midrule
			A  & \cmark & - & - & \cmark & $E_X$ to $E_Z$ & 71.4 &	80.6 &	67.8 
			\\
			\midrule
			\multirow{2}{*} B  & \multirow{2}{*} \cmark & \multirow{2}{*} \cmark & \multirow{2}{*} - & \multirow{2}{*} \cmark & $E_X$ to $E_Z$, &  \multirow{2}{*} {71.3} & \multirow{2}{*} {80.8} &	\multirow{2}{*} {68.4}
			\\
			& & & & & $E_X$ to $E_{DT}$ & & & \\
			\midrule
			\multirow{4}{*}C & \multirow{4}{*} \cmark & \multirow{4}{*} \cmark & \multirow{4}{*} \cmark & \multirow{4}{*} \cmark & $E_X$ to $E_Z$, &  \multirow{4}{*} {\textbf{73.2}} &	\multirow{4}{*} {\textbf{84.1}} &	\multirow{4}{*}{\textbf{69.9}}\\
			
			& & & & & $E_X$ to $E_{DT}$, & & & \\
			& & & & & $E_X$ to $E_{DB}$, & & & \\
			& & & & & $E_{DB}$ to $E_Z$ & & & \\
			\midrule
		\end{tabular}
	\end{center}
\end{table}

\subsubsection{Effectiveness of Search Token Partitioning in Deeper  Layers}\label{sec3subsec2subsse} 
\begin{figure}[t]%
	\centering
	\includegraphics[width=0.48\textwidth]{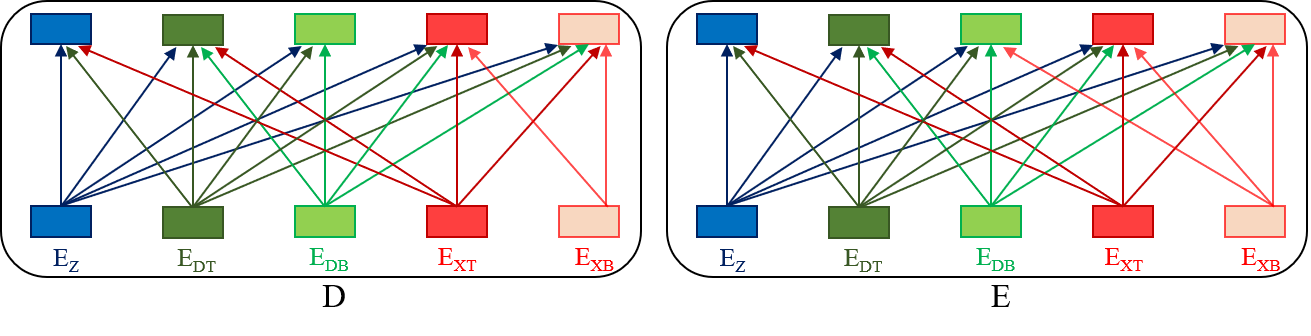}
	\caption{Information flow diagram for the deeper encoder layers of Model D and Model E. }
	\label{fig:Ablation3}
\end{figure}

In the deeper encoder layers of the proposed tracker, search region tokens are partitioned into target search tokens and non-target search tokens. To demonstrate the effectiveness of search token partitioning in deeper layers and to justify the interaction between token groups, we trained two models, referred to as Model D and Model E. The early layers of these two models are identical to those of Model C, and their information flow in deeper layers is illustrated in Fig. \ref{fig:Ablation3}. In both of these models, we blocked the information flow from non-target tokens to both initial target and dynamic target tokens as they reduce the importance of target-specific feature extraction.

\begin{table*}[t]
	\footnotesize
	\begin{center}
		\caption{Study on information flow among initial template ($E_Z$), dynamic template ($E_{DT}$), dynamic background ($E_{DB}$), target search ($E_{XT}$), and non-target search ($E_{XB}$) tokens in deeper encoder layers. }\label{tabel:Ablation3}%
		\begin{tabular}{ccccccccccc}
			\toprule
			\multirow{2}{*}{\textbf{Model}}& \multicolumn{6}{c}{\textbf{Token Groups}} & \multirow{2}{*}{\textbf{Blocked}}& \multicolumn{3}{c}{\textbf{GOT-10k}}\\ 			\cmidrule{2-7} \cmidrule{9-11}
			
			& $E_Z$ & $E_{DT}$ & $E_{DB}$ & $E_X$ & $E_{XT}$ & $E_{XB}$ & \textbf{Information Flow}  & AO & $\mathrm{SR_{0.5}}$ & $\mathrm{SR_{0.75}}$ \\	
			\midrule
			\multirow{2}{*} C & \multirow{2}{*} \cmark & \multirow{2}{*} \cmark & \multirow{2}{*} \cmark &  \multirow{2}{*} \cmark & \multirow{2}{*} - & \multirow{2}{*} - &  $E_X$ to $E_Z$, $E_X$ to $E_{DT}$, &  \multirow{2}{*} {73.2} &	\multirow{2}{*}{84.1} &	\multirow{2}{*}{69.9}
			\\
			& & & & & & & $E_X$ to $E_{DB}$, $E_{DB}$ to $E_Z$ & & & \\
			\midrule
			\multirow{3}{*} D & \multirow{3}{*} \cmark & \multirow{3}{*} \cmark & \multirow{3}{*} \cmark & \multirow{3}{*} - & \multirow{3}{*} \cmark & \multirow{3}{*} \cmark &  $E_{XB}$ to $E_Z$, $E_{XB}$ to $E_{DT}$, &  \multirow{3}{*}{\textbf{74.6}} &	\multirow{3}{*}{\textbf{85.6}} &	\multirow{3}{*}{\textbf{71.9}}
			\\
			& & & & & & & $E_{XB}$ to $E_{DB}$, $E_{DB}$ to $E_Z$, & & & \\
			& & & & & & &  $E_{XT}$ to $E_{DB}$ & & & \\
			\midrule
			\multirow{2}{*} E & \multirow{2}{*} \cmark & \multirow{2}{*} \cmark & \multirow{2}{*} \cmark & \multirow{2}{*} - & \multirow{2}{*} \cmark & \multirow{2}{*} \cmark &  $E_{XB}$ to $E_Z$, $E_{XB}$ to $E_{DT}$, &  \multirow{2}{*}{73.2} & \multirow{2}{*}{84.3} & \multirow{2}{*}{70.2} 
			\\
			& & & & & & & $E_{DB}$ to $E_Z$, $E_{XT}$ to $E_{DB}$ & & & \\
			\midrule
		\end{tabular}
	\end{center}
\end{table*}
From the experimental results on the GOT-10k dataset, as summarized in Table \ref{tabel:Ablation3}, it is clearly observed that partitioning search region tokens in deeper layers and then allowing the interaction from target search tokens to initial and dynamic target tokens enhances the tracking performances since both Models D and E showed better performance than Model C. It is also observed that allowing the information flow from non-target search tokens to dynamic background tokens reduces tracking performance, as the background cues from the large number of non-target tokens distract the dynamic background cues. This is evident in the poorer tracking performance of Model E compared to Model D.

\begin{table}[!t]
	\footnotesize
	\begin{center}
		\caption{Ablative experiments on the number of target search tokens in deeper encoder layers.}\label{tabel:Ablation4}%
		\begin{tabular}{lllll}
			\toprule
			\multirow{2}{*}{\textbf{Number of Tokens}}& \multicolumn{3}{c}{\textbf{GOT-10k}}\\ 
			\cmidrule{2-4}
			&  AO & $\mathrm{SR_{0.5}}$ & $\mathrm{SR_{0.75}}$ \\
			\midrule
			20 Tokens & 73.5 &	84.5 &	71.0 \\
			40 Tokens & 73.6 &	84.1 &	71.2 \\
			64 Tokens & \textbf{74.6} &	\textbf{85.6} & \textbf{71.9} \\
			89 Tokens & 73.4 &	83.9 & 70.2 \\
			
			\midrule
		\end{tabular}
	\end{center}
\end{table}
At the onset of deeper encoder layers, we choose K search tokens as target search tokens, determined by their attention score weights as outlined in Equation \ref{eqn:token}. We have selected 64 target search tokens ($K=64$) based on experimental results. The summary of experimental results used to determine K is provided in Table \ref{tabel:Ablation4}.

\subsection{Experimental Results and Comparisons}\label{sec3subsec3}

\begin{table*}
	\begin{center}
		\caption{Tracking the performance comparison of the proposed OIFTrack approach with state-of-the-art trackers on GOT-10k, TrackingNet, and LaSOT benchmarks. The abbreviations are denoted as $\mathrm{AO}$ for Average Overlap, $\mathrm{SR_{0.5}}$ for Success Rate at the $0.5$ threshold, $\mathrm{SR_{0.75}}$ for Success Rate at the $0.75$ threshold, $\mathrm{AUC}$ for Area Under the Curve (AUC) of success between 0 and 1, $\mathrm{P}$ for Precision, and $\mathrm{P_n}$ for Normalized Precision. All results are reported in percentages. The \textcolor{red}{red}, \textcolor{blue}{blue}, and \textcolor{ForestGreen}{green} colors are used to respectively denote the top three results. The $\ast$ symbol is used to indicate that we only consider trackers trained exclusively on the GOT-10k training set.  }\label{MasterTable}%
		\begin{tabular}{lcl@{\extracolsep{0.30cm}}l@{\extracolsep{0.30cm}}l@{\extracolsep{0.8cm}}l@{\extracolsep{0.30cm}}l@{\extracolsep{0.30cm}}l@{\extracolsep{0.8cm}}l@{\extracolsep{0.30cm}}l@{\extracolsep{0.30cm}}l}
			\midrule
			\multirow{2}{*}{\textbf{Trackers}} & 
			\multirow{2}{*}{\textbf{Source}} & \multicolumn{3}{c}{\textbf{GOT-10k $\ast$}} &
			\multicolumn{3}{c}{\textbf{TrackingNet}} & \multicolumn{3}{c}{\textbf{LaSOT}}\\ 
			\cmidrule{3-5} \cmidrule{6-8} \cmidrule{9-11}
			& & AO & $\mathrm{SR_{0.5}}$ & $\mathrm{SR_{0.75}}$ & AUC & $\mathrm{P_{n}}$ & P &  AUC & $\mathrm{P_{n}}$ & P  \\
			\midrule
			OIFTrack & Ours & \color{red}74.6 & \color{red}85.6 & \color{red}71.9 & \color{red}84.1 & \color{red}89.0 & \color{blue}82.8 & \color{blue}69.6 & \color{red}79.5 & \color{ForestGreen}75.4 
			\\
			SMAT \citep{Gopal_2024_WACV} & WACV'24 & 64.5 & 74.7 & 57.8 & 78.6 & 84.2 & 75.6 & 61.7 & 71.1 & 64.6 
			\\
			GRM \citep{gao2023generalized} & CVPR'23 & \color{blue}73.4 &	\color{ForestGreen}82.9 &	\color{blue}70.4	& \color{blue}84.0	& \color{blue}88.7 & \color{red}83.3  & \color{red}69.9 & \color{ForestGreen}79.3 & \color{red}75.8 
			\\
			F-BDMTrack-256 \citep{Yang_2023_ICCV} & ICCV'23 & 72.7 & 82.0 & 69.9 & \color{ForestGreen}83.7 & 88.3 & 82.6 & \color{red}69.9 & \color{blue}79.4 & \color{red}75.8 \\
			TATrack-B \citep{He_Zhang_Xie_Li_Wang_2023} & AAAI'23 & \color{ForestGreen}73.0 & \color{blue}83.3 & 68.5 & 83.5 & 88.3 & 81.8 & \color{ForestGreen}69.4 & 78.2 & 74.1 
			\\
			ROMTrack \citep{Cai_2023_ICCV} & ICCV'23 & 72.9 & \color{ForestGreen}82.9 & 70.2 & 83.6 & \color{ForestGreen}88.4 & \color{ForestGreen}82.7 & 69.3 & 78.8 & \color{ForestGreen}75.6
			\\	
			CTTrack-B \citep{Song_Luo_Yu_Chen_Yang_2023} & AAAI'23 & 71.3 & 80.7 & \color{ForestGreen} 70.3 & 82.5 & 87.1 & 80.3 & 67.8 & 77.8 & 74.0
			\\
			MATTrack \citep{Zhao_2023_CVPR} & CVPR'23 & 67.7	 & 78.4	 & -	& 81.9 & 86.8  & - & 67.8 & 77.3 &  - 
			\\
			GdaTFT \citep{Liang_Li_Long_2023} & AAAI'23 & 65.0 &  77.8 & 53.7 & 77.8 & 83.5 & 75.4 & 64.3 & 68.0 & 68.7 \\			
			BANDT \citep{Yang_2023_IEEETrans} & IEEE TCE'23 & 64.5 & 73.8 & 54.2  & 78.5 & 82.7 & 74.5 & 64.4 & 72.4 & 67.0
			\\
			HiT-Base \citep{Kang_2023_ICCV} & ICCV'23 & 64.0 & 72.1 & 58.1 & 80.0 & 84.4 & 77.3 & 64.6 & 73.3 & 68.1 
			\\
			OSTrack-256 \citep{ye2022joint} & ECCV'22 & 71.0 &	80.4 &	68.2 & 83.1	& 87.8	& 82.0  & 69.1 & 78.7 & 75.2
			\\
			MixFormer-22k \citep{cui2022mixformer} & CVPR'22 & 70.7 &	80.0 & 67.8	 &	83.1 &	88.1 &	81.6 &	69.2 &	78.7 &	74.7	
			\\
			SimTrack-B/16 \citep{chen2022backbone} & ECCV'22 &	69.8 &	78.8 &	66.0  &	82.3 &	86.5 &	80.2 & 	69.3 &	78.5 &	74.0
			\\
			AiATrack \citep{gao2022aiatrack} & ECCV'22 & 69.6 &	80.0 &	63.2 &	82.7 &	87.8 & 80.4 	&	69.0 &	\color{blue}79.4 &	73.8							
			\\
			CSWinTT \citep{song2022transformer} & CVPR'22 &	69.4 &	78.9 &	65.4 &	81.9 &	86.7 &	79.5 &	66.2 &	75.2 &	70.9
			\\
			SwinTrack-B\citep{lin2022swintrack} & NeurlPS'22 & 69.4 &	78.0 &	64.3 & 82.5 & 87.0 & 80.4 &	\color{blue}69.6 &	78.6 &	74.1			
			\\
			SparseTT \citep{fu2022sparsett} & IJCAI'22 & 69.3 &	79.1 &	63.8 &	81.7 &	86.6 &	79.5 &	66.0 &	74.8 &	70.1				
			\\
			ToMP \citep{mayer2022transforming} & CVPR'22 &	- &	- &	-  &	81.5 &	86.4 &	78.9 & 68.5 & 79.2 &	73.5
			\\
			DTT \citep{yu2021high} & ICCV'21 & 68.9 & 79.8 & 62.2 & 79.6 & 85.0 & 78.9 & 60.1 & - & -
			\\	
			STARK \citep{yan2021learning} & ICCV'21 & 68.8 &	78.1 &	64.1 &	82.0 &	86.9 &	79.1 &	67.1 &	77.0 &	72.2
			\\			
			TransT \citep{chen2021transformer} & CVPR'21 & 67.1 &	76.8 &	60.9  &	81.4 &	86.7 &	80.3  &	64.9 &	73.8 &	69.0
			\\
			TrDiMP \citep{wang2021transformer} & CVPR'21 &	67.1 &	77.7 &	58.3 &	78.4 &	83.3 &	73.1 &	63.9 &	73.0 &	61.4
			\\
			SiamGAT \citep{guo2021graph} & CVPR'21 &	62.7 &	74.3 &	48.8 &	- &	- &	-  & 53.9 &	63.3 &	53.0
			\\			
			SiamR-CNN \citep{voigtlaender2020siam} & CVPR'20 &	64.9 &	72.2 &	68.4 &	81.2 &	85.4 &	80.0  &	64.8 &	72.2 &	60.8
			\\
			PrDiMP \citep{danelljan2020probabilistic} & CVPR'20 & 63.4 &	73.8 &	54.3 &	75.8 &	81.6 &	70.4 & 59.8 &	- &	-			
			\\
			Ocean \citep{zhang2020ocean} & ECCV'20 &	61.1 &	72.1 &	47.3 &	- & - &	- & 56.0 &	65.1 & 56.6						
			\\
			DiMP \citep{Bhat_2019_ICCV} & ICCV'19 & 61.1 &	71.7 &	49.2  &	74.0 &	80.1 &	68.7 & 56.9 &	- & -						
			\\			
			SiamRPN++ \citep{li2019siamrpn++} & CVPR'19 & - &	- &	- &	73.3 &	80.0 &	69.4 & 49.6 &	56.9 &	49.1			
			\\		
			\bottomrule
		\end{tabular}
	\end{center}
	
\end{table*}
The tracking performance of the proposed OIFTrack approach was evaluated on the GOT-10K \citep{huang2019got}, LaSOT \citep{Fan_2019_CVPR}, TrackingNet \citep{Muller_2018_ECCV}, and UAV123 \citep{UAV123_2016} benchmarks and the reported results of the state-of-the-art trackers are used for comparison. We have compared the performances of the proposed tracker with Transformer-based trackers:   GRM \citep{gao2023generalized}, TATrack \citep{He_Zhang_Xie_Li_Wang_2023}, ROMTrack \citep{Cai_2023_ICCV}, CTTrack \citep{Song_Luo_Yu_Chen_Yang_2023}, MATTrack \citep{Zhao_2023_CVPR}, HiT \citep{Kang_2023_ICCV}, OSTrack \citep{ye2022joint}, MixFormer \citep{cui2022mixformer}, SimTrack \citep{chen2022backbone}, SwinTrack \citep{lin2022swintrack}, and SparseTT \citep{fu2022sparsett} CNN-Transformer based trackers: SMAT \citep{Gopal_2024_WACV}, BANDT \citep{Yang_2023_IEEETrans}, AiATrack \citep{gao2022aiatrack}, CSWinTT \citep{song2022transformer}, ToMP \citep{mayer2022transforming}, DTT \citep{yu2021high}, STARK \citep{yan2021learning}, TransT \citep{chen2021transformer}, and TrDiMP \citep{wang2021transformer}, and CNN-based trackers: GdaTFT \citep{Liang_Li_Long_2023}, SiamGAT \citep{guo2021graph}, SiamR-CNN \citep{voigtlaender2020siam}, PrDiMP \citep{danelljan2020probabilistic}, Ocean \citep{zhang2020ocean}, DiMP \citep{Bhat_2019_ICCV}, and SiamRPN++ \citep{li2019siamrpn++}. While some trackers trained and evaluated their models on both small and large search region sizes of $256 \times 256$ and $384 \times 384$ respectively, we only consider the reported results of $256 \times 256$ search region models to compare performances without bias. The tracking robustness comparison of the proposed tracker for the GOT-10K, LaSOT, and TrackingNet datasets is provided in Table \ref{MasterTable}.

\subsubsection{GOT-10k}\label{sec3subsec3sec1}
We evaluate our tracker using GOT-10k to assess its one-shot tracking capability, as the training object classes in GOT-10k do not overlap with the testing object classes. Also, the evaluation results of this dataset remain unbiased toward familiar objects, as tracking models exclusively train on the GOT-10k training set. We strictly adhere to GOT-10K protocols for training, and the proposed tracker's performance is measured based on evaluation results obtained from the dataset's official evaluation server.

Experimental results in Table \ref{MasterTable} clearly indicate the superior performance of the proposed OIFTrack approach, as it outperforms recent one-stream Transformer trackers and other competitors by a considerable margin. Our tracker achieved a 74.6\% AO score, which is a 1.2\% improvement compared to the second-top performing tracker, GRM \citep{gao2023generalized}. Similarly, the proposed OIFTrack demonstrated a 2.3\% and 1.5\% improvement over the second-top performing trackers in terms of success rate at 0.5 threshold and 0.75 threshold, respectively. The outstanding performance of the proposed tracker on the GOT-10K benchmark illustrates its one-stream tracking capability by extracting more discriminative features for unseen object classes.

\subsubsection{TrackingNet}\label{sec3subsec3sec3}
The TrackingNet dataset has a massive number of tracking sequences with a large diversity of target object classes, varying resolutions, and frame rates. Based on the experimental results, our tracker demonstrated better performance than state-of-the-art trackers, achieving a top AUC score of 84.1\% and a normalized precision score of 89.0\%, while also achieving the second-best precision score of 82.8\%. The superior performance of OIFTrack on the TrackingNet dataset demonstrates its robustness in real-world tracking capabilities.

\subsubsection{LaSOT}\label{sec3subsec3sec2}
The LaSOT benchmark dataset contains a large-scale collection of long-term tracking sequences obtained from various challenging scenarios. The average length of a test sequence is almost 2500 frames, and the dataset contains 280 sequences in the test set. Based on the experimental results Table in \ref{MasterTable}, our OIFTrack approach showed competitive performances in the LaSOT benchmark with a top normalized precision score of 79.5\%, the second-highest AUC score of 69.6\%, and the third-highest precision score of 75.4\%. In addition to the overall performance evaluation, we have conducted attribute-wise performance measures on LaSOT, and the success plots are shown in Fig. \ref{Attribute_analysis_LaSOT}. The competitive performance of the OIFTrack approach on LaSOT showed its long-term tracking capabilities in challenging scenarios.

\subsubsection{UAV123}\label{sec3subsec4sec4}
We evaluate the performance of our tracker on the UAV123 dataset to determine its effectiveness in aerial tracking. The UAV123 dataset \citep{UAV123_2016} comprises 123 video sequences captured by unmanned aerial vehicles (UAVs). Tracking targets in the UAV123 dataset presents a challenge due to the relatively small size of target objects in aerial tracking sequences, coupled with frequent changes in both the target object's position and the camera's orientation. As reported in Table \ref{comparison_UAV123}, our tracker achieved competitive results, showing comparability with recent trackers like OSTrack-256 \citep{ye2022joint} and ARTrack-256 \citep{Wei2023autoregressive}. Specifically, our tracker attained an AUC score of 68.6\% and a Precision of 90.3\%.

\begin{table}[!t]
	\footnotesize
	\begin{center}
		\caption{Tracking the performance comparison of the proposed OIFTrack approach with state-of-the-art trackers on UAV123. The abbreviations are denoted as $\mathrm{AUC}$: Area Under the Curve (AUC) of success between 0 to 1, and $\mathrm{P}$:Precision. All the results are reported in percentage. The best two results are shown in \textcolor{red} {red} and \textcolor{blue} {blue} colors, respectively.}\label{comparison_UAV123}%
		\begin{tabular}{llll}
			\midrule
			\multirow{2}{*}{\textbf{Trackers}} & \multirow{2}{*} \textbf{Year}& \multicolumn{2}{c}{\textbf{UAV123}}\\ 
			\cmidrule{3-4}
			& & AUC & P \\
			\midrule			
			SMAT \citep{Gopal_2024_WACV} & 2024 & 64.3 & 83.9
			\\
			HiT-Base \citep{Kang_2023_ICCV} & 2023 & 65.6 & -
			\\
			ARTrack-256 \citep{Wei2023autoregressive} &2023 & 67.7 & -
			\\
			MATTrack \citep{Zhao_2023_CVPR} & 2023 & 68.0 & -
			\\
			TransT \citep{chen2021transformer} & 2021 &	68.1 & 87.6
			\\
			OSTrack-256 \citep{ye2022joint} & 	2022 &  68.3 & 88.8
			\\
			STARK \citep{yan2021learning} & 2021 &	68.5 & 89.5
			\\
			CTTrack-B \citep{Song_Luo_Yu_Chen_Yang_2023} & 2023 & \color{red}68.8 & \color{blue}89.5
			\\
			OIFTrack &  & \color{blue}68.6 & \color{red}90.3\\				
			\midrule
		\end{tabular}
	\end{center}
\end{table}

\subsection{Discussion}
\begin{figure*}[t]%
	\centering
	\includegraphics[width=0.98\textwidth]{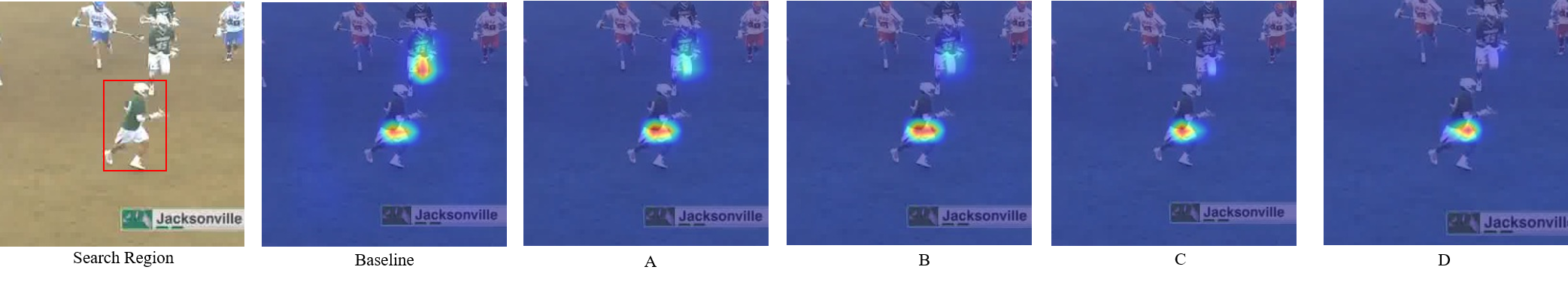}
	\caption{Illustration of the effectiveness of the OIFTrack approach through step-by-step visualization results of classification heatmaps with various information flows. Baseline: free bidirectional flow; Model A: blocks the flow from search to target tokens; Model B: includes dynamic target cues; Model C: includes dynamic background cues; Model D: allows interaction between target search tokens and other target tokens in deeper layers.}
	\label{fig:Ablation}
\end{figure*}

\begin{figure}[t]%
	\centering
	\includegraphics[width=0.5\textwidth]{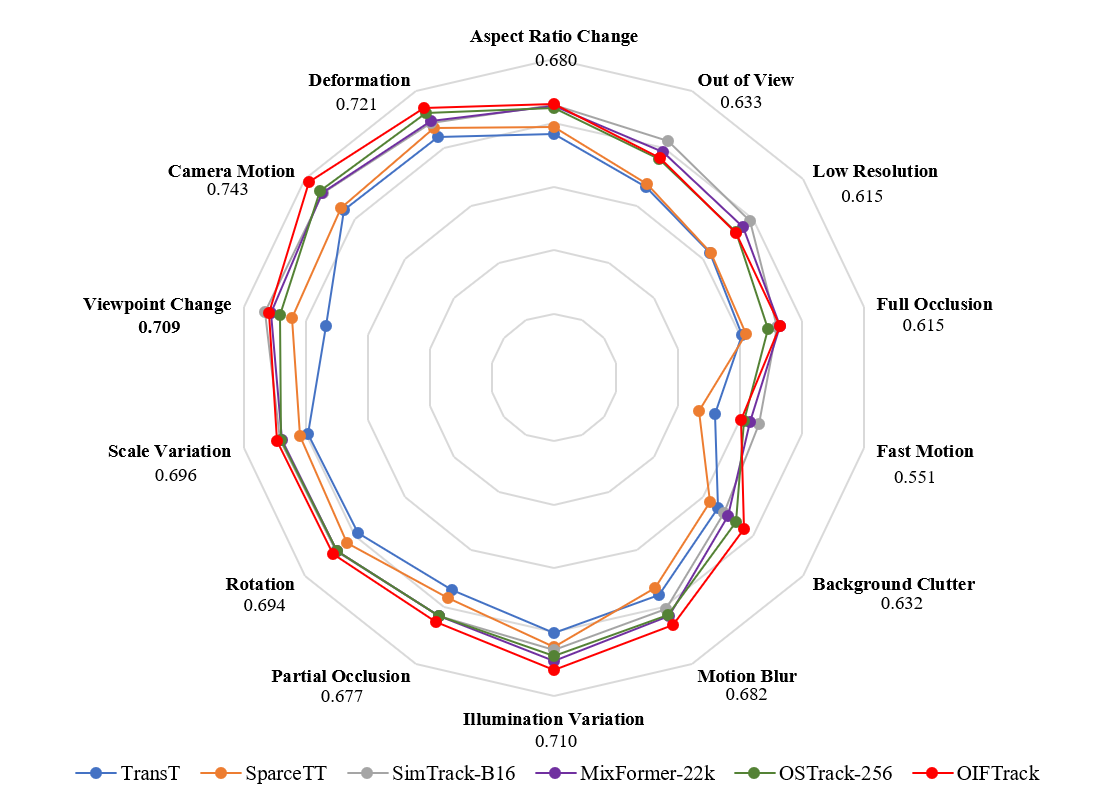}
	\caption{AUC score comparison of State-of-the-art trackers on LaSOT for 14 challenging attributes}
	\label{Attribute_analysis_LaSOT}
\end{figure}
In this study, we propose a novel one-stream Transformer tracker that blocks unnecessary information flow between tokens, rather than relying on the free bidirectional information flow used in previous approaches. In addition, we have utilized dynamic temporal cues to capture surrounding information about the target, which is then leveraged to mitigate the influence of distractor objects in the search region. To demonstrate the effectiveness of the proposed OIFTrack approach, the heatmap of the classification scores for a search region is visualized and shown in Fig. \ref{fig:Ablation}. Based on the figure, it is evident that in the heatmap of the baseline tracker, the target and similar distractor objects have equal scores, as the information flow from a large number of background tokens in search region distracts the target-specific features.  Although the classification scores of the distractor are reduced in the heatmap of Model A by blocking unnecessary flow from search to target, they are entirely diminished in the heatmap of Model C after capturing the surrounding information of the target from dynamic background cues. Furthermore, the classification scores of the target became stronger in Model D after allowing the interaction of target search tokens with other tokens that have target cues. 

In addition to the overall performance, our tracker demonstrated better performance in many tracking scenarios compared to other Transformer-based trackers, as shown in Fig. \ref{Attribute_analysis_LaSOT} through attribute-wise comparison. Since our OIFTrack approach captures the appearance changes of the target from the interaction between the initial target, dynamic target, and target search tokens, it exhibited superior performance in rotation, scale variation, deformation, as well as full and partial occlusion scenarios compared to other one-stream Transformer trackers. Furthermore, our tracker performs well in background clutter and camera motion scenarios due to its ability to capture the surrounding information of the target.

Our OIFTrack runs at an average speed of 21 FPS on a single core of a Tesla P100 GPU and it contains 92.63 million parameters and 33.82 giga floating point operations (FLOPS). Based on computational efficiency, our tracker is slightly lower than similar approaches such as OSTrack and GRM, due to our incorporation of dynamic target and background cues into the tracking process. As another limitation, although our tracker showed better performance in long-term tracking according to the results on the LaSOT benchmark, the difference between our performance and that of competitors is marginal. Since the probability of wrong identification of the dynamic template in long-term sequences is high, that may be the reason for this performance degradation.

\section{Conclusion}\label{sec5}
In this work, we present a novel one-stream Transformer tracker aimed at enhancing target-specific feature extraction and utilizing  surrounding information of the target.  We have analyzed the information flow between tokens and then removed the unnecessary flows using a simple attention masking mechanism. In addition, we utilized dynamic background cues to capture the surrounding information of the target, thereby reducing the impact of distractor objects. Through extensive experiments on popular benchmark datasets, the proposed OIFTrack approach demonstrated outstanding performance, especially in one-shot tracking.

\section*{CRediT authorship contribution statement}
\textbf{Janani Kugarajeevan}: Conceptualization, Methodology,  Investigation, Implementation, Validation, Writing - Original Draft, Visualization. \textbf{Kokul Thanikasalam}: Conceptualization, Investigation, Methodology, Supervision, Writing - Original Draft. \textbf{Amirthalingam Ramanan}: Conceptualization, Supervision, Writing - Review \& Editing. \textbf{Subha Fernando}: Conceptualization, Supervision, Writing - Review \& Editing

\section*{Declaration of competing interest}
The authors declare that they have no known competing financial interests or personal relationships that could have appeared to influence the work reported in this paper.

\section*{Data availability}
Following publicly available benchmark datasets are used in this study. 
\begin{itemize}
	\itemsep0em 
	\item GOT-10k: \url{http://got-10k.aitestunion.com}
	\item TrackingNet: \url{https://tracking-net.org}
	\item LaSOT: \url{http://vision.cs.stonybrook.edu/~lasot}
	\item UAV123: \url{https://cemse.kaust.edu.sa/ivul/uav123}
\end{itemize}
\bibliographystyle{model5-names} 
\bibliography{sn-bibliography}

\end{document}